\documentclass[letterpaper, 10 pt, conference]{ieeeconf}  

\IEEEoverridecommandlockouts                              

\usepackage{amsmath,amsfonts}
\usepackage[ruled,vlined]{algorithm2e}
\usepackage{algpseudocode}
\usepackage{array}
\usepackage{textcomp}
\usepackage{stfloats}
\usepackage{url}
\usepackage{verbatim}
\usepackage{graphicx}
\def\BibTeX{{\rm B\kern-.05em{\sc i\kern-.025em b}\kern-.08em
    T\kern-.1667em\lower.7ex\hbox{E}\kern-.125emX}}
\usepackage{balance}
\usepackage{graphics}
\usepackage{subfigure}
\usepackage{booktabs}
\usepackage{multirow}
\usepackage{cite}
\usepackage{tikz,xcolor,hyperref}

\usepackage{fontawesome}
\usepackage[top=54pt,bottom=54pt, left=54pt, right=54pt]{geometry}

\vspace{18pt}
\title{\LARGE \bf
Observation Time Difference: an Online Dynamic Objects Removal Method for Ground Vehicles
}

\author{Rongguang Wu, Chenglin Pang, Xuankang Wu, Zheng Fang${}^*$
\thanks{This work was supported in part by the National Natural Science Foundation of China under Grants 62073066 and U20A20197, in part by the Fundamental Research Funds for the Central Universities under Grant N2226001, and in part by 111 Project under Grant B16009 (Corresponding author: Zheng Fang). }
\thanks{The authors are all with the Faculty of Robot Science and Engineering, Northeastern University, Shenyang 110819, China (e-mail: 2202066@stu.neu.edu.cn, 2010690@stu.neu.edu.cn, 2202067@stu.neu.edu.cn, fangzheng@mail.neu.edu.cn).}
}

\begin{document}

\maketitle
\thispagestyle{empty}
\pagestyle{empty}

\begin{abstract}
In the process of urban environment mapping, the sequential accumulations of dynamic objects will leave a large number of traces in the map. These traces will usually have bad influences on the localization accuracy and navigation performance of the robot. Therefore, dynamic objects removal plays an important role for creating clean map. However, conventional dynamic objects removal methods usually run offline. That is, the map is reprocessed after it is constructed, which undoubtedly increases additional time costs. To tackle the problem, this paper proposes a novel method for online dynamic objects removal for ground vehicles. According to the observation time difference between the object and the ground where it is located, dynamic objects are classified into two types: \emph{suddenly appear} and \emph{suddenly disappear}. For these two kinds of dynamic objects, we propose downward retrieval and upward retrieval methods to eliminate them respectively. We validate our method on SemanticKITTI dataset and author-collected dataset with highly dynamic objects. Compared with other state-of-the-art methods, our method is more efficient and robust, and reduces the running time per frame by more than 60$\%$ on average. Our method will be open-sourced on GitHub\footnote{https://github.com/RongguangWu/OTD}.

\end{abstract}

\section{Introduction}
\label{sec:introduction}

\begin{figure}[htbp]
\centering
\includegraphics[width=\columnwidth]{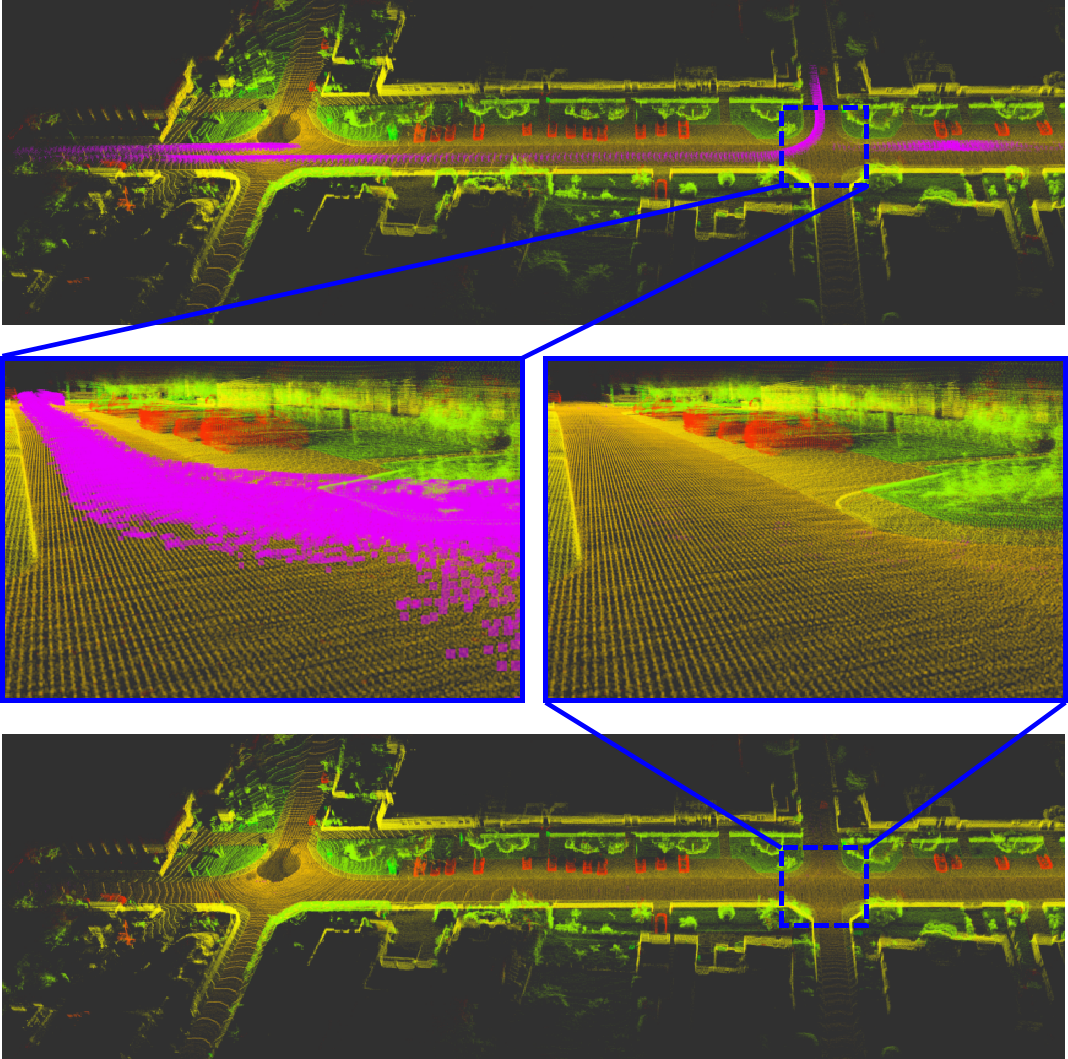}
\caption{The results of dynamic objects removal. The upper image is original map. The lower image is static map constructed by our method. The middle image is partially enlarged image, and the purple area is traces left by dynamic objects.}
\label{fig::firstpage}
\end{figure}

Concise and reliable maps are the basis for long-term autonomy of mobile robots\cite{kim20191, stefanini2022efficient}. With the information retrieved from the map, robots can achieve successful localization and navigation\cite{adkins2022probabilistic, wang2022ltsr}. Currently, real-time mapping using SLAM-based technology, especially LiDAR mapping, has become a common way to obtain maps due to many advantages\cite{xu2022fast, shan2020lio}. Unfortunately, there are inevitably a substantial amount of dynamic objects\cite{pagad2020robust, pomerleau2014long} during urban mapping, which will leave undesired traces when constructing the point cloud maps\cite{lim2021erasor}. Moreover, these traces will form a huge impassable area (as shown in Fig. \ref{fig::firstpage}), which will make subsequent tasks difficult, such as navigation and path planning. Therefore, dynamic objects removal is crucial in constructing clean point cloud map.

Existing dynamic objects removal methods can be mainly divided into two categories: offline removal and online removal. Offline removal is to perform dynamic removal on the basis of a pre-built map (prior map), in order to obtain a new map with only static objects. Several effective methods have been proposed, such as \cite{schauer2018peopleremover, kim2020remove, lim2021erasor}. Although these methods achieve significant dynamic removal performance, they require additional time cost. Therefore, this paper focuses on removing dynamic objects while building the map, namely online dynamic objects removal. 
 
For online dynamic objects removal, an effective idea is to apply and improve similar ideas used in offline removal methods, such as \cite{park2022nonparametric, fan2022dynamicfilter}. But the core principle of offline removal method is to judge dynamic objects by comparing the difference between the point cloud of each frame and the prior map. It is infeasible to directly apply offline methods to online removal, since online removal does not have a prior map. Even though online map can be obtained by SLAM algorithms, the map only contains information prior to current frame, and no information after current frame. This will lead to a decrease in the performance of dynamic removal.

In this paper, we propose a novel online dynamic objects removal method for ground vehicles. Considering that common dynamic objects (vehicles, pedestrians, etc.) are always in contact with the ground\cite{lim2021erasor}, our method judges the dynamic objects by comparing the observation time difference between the object and the ground on which it is located. Our contributions are as follows:

\begin{itemize}
\item Based on the observation time difference between the object and the ground it is located on, we classify dynamic objects into two categories: \emph{suddenly appear} and \emph{suddenly disappear}.

\item For the dynamic objects of \emph{suddenly appear} and \emph{suddenly disappear}, we propose two methods, namely downward retrieval and upward retrieval, to remove them respectively.

\item We conduct extensive experimental validations on challenging datasets, and SemanticKITTI\cite{behley2019semantickitti} datasets. Compared to the state-of-the-art (SOTA) methods, our approach is more effective and robust, and reduces the average processing time per frame by over 60$\%$.

\item We have encapsulate our proposed method as an independent dynamic removal module, which can run concurrently with LiDAR mapping algorithm to generate clean static maps. Our method will be open-sourced on GitHub.
\end{itemize}

The rest of the paper is organized as follows: Section \ref{sec::related-work} reviews the related work about dynamic objects removal. In section \ref{sec::classification}, we divide dynamic objects into two categories. Section \ref{sec::methodology} explains the details of our proposed dynamic removal method. Experimental comparisons and analyses are carried out in Section \ref{sec::experiments}. Finally, Section \ref{sec::conclusion} summarizes the contributions and describes future works.

\section{Related Work}
\label{sec::related-work}
This section briefly reviews the relevant studies on dynamic objects removal methods, which can be roughly divided into offline removal approaches and online removal approaches:

\subsection{Offline Removal Approaches}
One mainstream concept of offline removal is to detect dynamic objects by comparing the differences between the point cloud of each frame and the prior map. Ray tracing-based method is a commonly used method. Its fundamental idea is to calculate the occupancy probability of a voxel based on whether it is penetrated by the laser beam of the LiDAR\cite{hornung2013octomap}. However, checking voxels is computationally expensive. Even though Schauer $et$ $al.$\cite{pomerleau2014long} and Pagad $et$ $al.$\cite{pagad2020robust} made some improvements to accelerate the speed of inspecting voxels, the processing time is still long. 

In order to reduce the computational cost, visibility-based method is proposed\cite{pomerleau2014long, kim2020remove}, which determines dynamic objects by checking whether the points in the map are occluded by the points in the query frame. However, when the incidence angle is big, the visibility-based method may mistakenly identify distant ground points as dynamic points. 

Additionally, Lim $et$ $al.$\cite{lim2021erasor, lim2023erasor2} proposed a visibility-free approach. This method is based on the assumption that dynamic objects are mostly in contact with the ground. It detects dynamic objects by comparing the pseudo-occupancy differences between the query frame and the map in the occupied grid. However, this method requires significant time consumption, making it unsuitable for real-time online processing.

\subsection{Online Removal Approaches}
In recent years, with the development of deep learning, numerous online dynamic objects segmentation methods have been proposed. Chen $et$ $al.$\cite{chen2021moving} generated residual images from previous scans and input them together with the range image of current scan to CNN, in order to distinguish between dynamic objects and static objects. Sun $et$ $al.$\cite{sun2022efficient} exploited  a dual-branch structure based range images, which is utilized to process the spatial-temporal information of LiDAR point cloud separately. Then, the motion-guided attention module was employed to detect dynamic objects. However, learning-based methods heavily rely on the training dataset, and they are prone to failure when there is a significant discrepancy between the actual scenario and the training scenario.

Furthermore, there are also some methods that employ a concept akin to offline removal\cite{park2022nonparametric, fan2022dynamicfilter}. But the problem is that online removal does not have a prior map. To solve this problem, Park $et$ $al.$\cite{park2022nonparametric} accumulate differential range images between the scan data of current frame and that of past few frames to obtain a background model. However, the background model only contain information prior to the current frame, and cannot capture information after current frame. Fan $et$ $al.$\cite{fan2022dynamicfilter} accumulated multi-frame point clouds to build a local submap, and then compared each frame with the local submap to judge dynamic objects. Nevertheless, the accumulated number of frames can impact the performance of dynamic removal. Too few frames may not provide sufficient information, while too many frames can increase the computational cost of the algorithm.

In this paper, we propose a novel online dynamic objects removal method. Our method determines dynamic objects by comparing observation time difference between the object and the ground, and achieves effective dynamic object removal without a prior map.

\section{Dynamic objects classification}
\label{sec::classification}
\begin{figure}[h]
\centering
\subfigure[\emph{suddenly appear} dynamic object]{
    \includegraphics[width=0.95\linewidth]{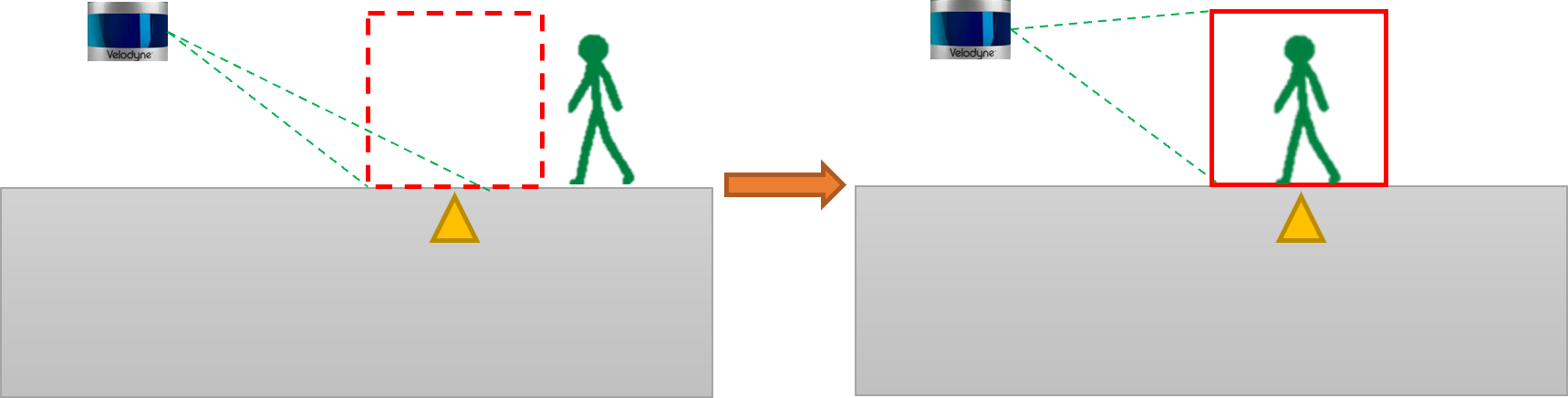}}

\subfigure[\emph{suddenly disappear} dynamic object]{
    \includegraphics[width=0.95\linewidth]{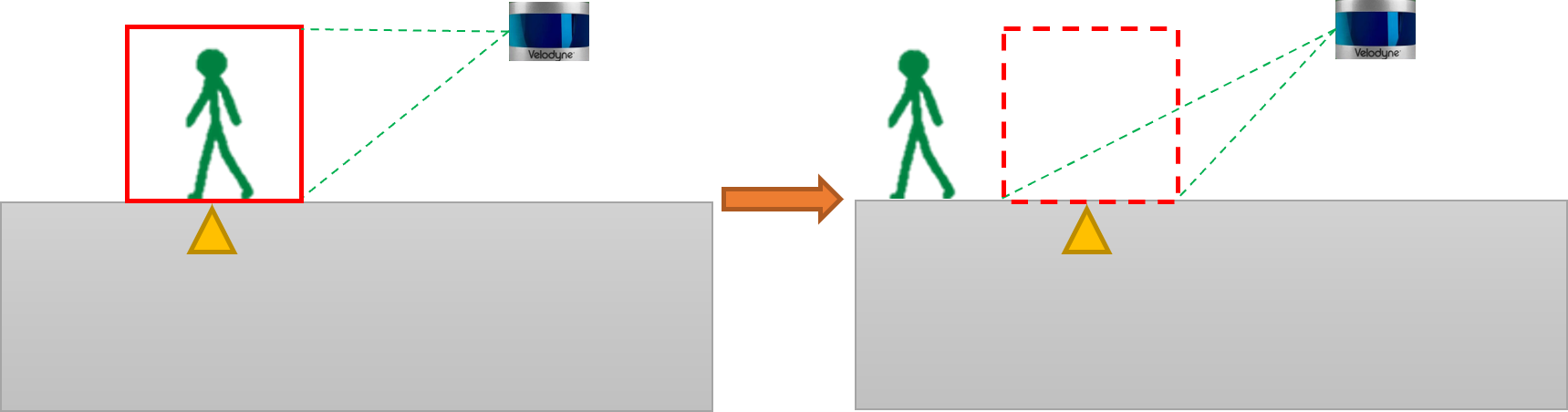}}
\caption{Dynamic objects classification. Only ground was observed at the begining in (a). But after a period of time, both ground and non-ground object were observed. The object is a \emph{suddenly appear} dynamic object. On the contrary, both ground and non-ground object were observed at the beginging in (b). But after a period of time, only ground was observed. The object is a \emph{suddenly disappear} dynamic object.}
\label{fig::dy_class}
\end{figure}

Similar to \cite{lim2021erasor}, we believe that in most cases, dynamic objects should have contact with the ground, whether they are pedestrians or terrestrial vehicles. Furthermore, if an object on the ground is static, it should be observed and disappear from the field of view simultaneously with the ground it is on. In other words, the object and the ground must appear and disappear simultaneously. If either of the two conditions is not met, the object is considered as a dynamic object. Therefore, we classify dynamic objects into the following two categories: 

\begin{itemize}
	\item As shown in Fig. \ref{fig::dy_class}(a), at a certain position, only ground was observed but not non-ground object at time ${t}_{0}$. Both ground and non-ground object were observed since time ${t}_{1}\left( {t}_{1} > {t}_{0} \right)$. Then, the non-ground object is defined as a \emph{suddenly appear} dynamic object.
 
	\item As shown in Fig. \ref{fig::dy_class}(b), at a certain position, both ground and non-ground object were observed at time ${t}_{0}$. Only ground was observed but not non-ground object since time ${t}_{1}\left( {t}_{1} > {t}_{0} \right)$. Then, the non-ground object is defined as a \emph{suddenly disappear} dynamic object.
\end{itemize}

\begin{figure*}[t]
\centering
\includegraphics[width=0.95\linewidth]{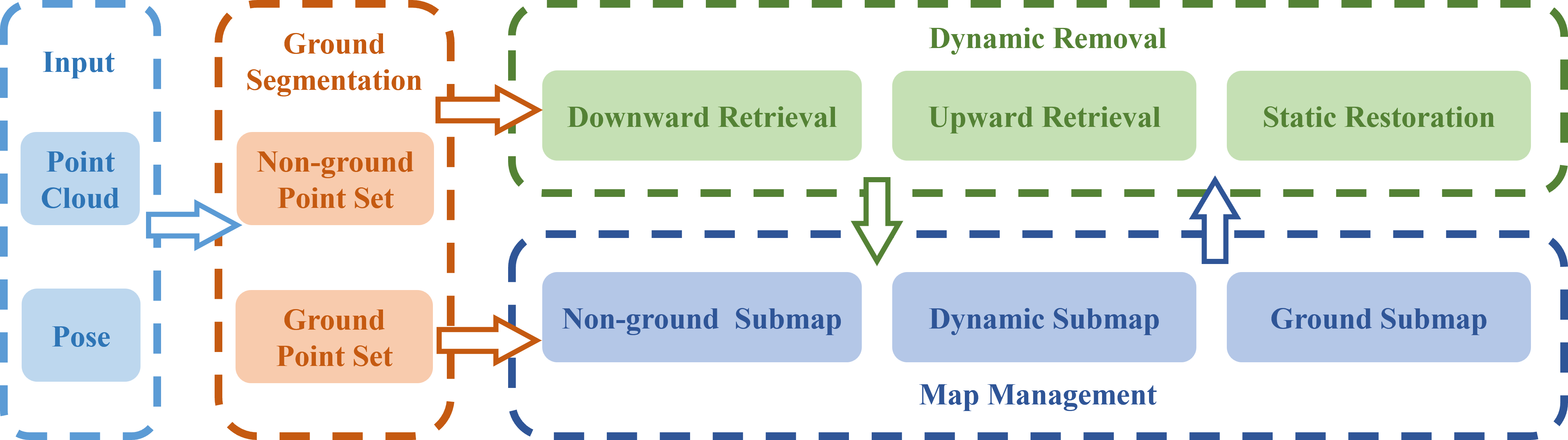}
\caption{Overview of our proposed method. Our proposed method includes three modules: ground segmentation, map management and dynamic removal. The point cloud $\mathbf{P}$ is divided into ground point set $\mathbf{G}$ and non-ground point set $\mathbf{U}$ in ground segmentation module, and then is sent to map management module and dynamic removal module together with the input pose $\mathbf{T}^W$. After downward retrieval, upward retrieval and static restoration, dynamic objects can be recognized.}
\label{fig::pipeline}
\vspace{-1em}
\end{figure*}

Note that a continuously moving object simultaneously meets the definitions of \emph{suddenly appear} and \emph{suddenly disappear}. When an object moves to a specific position, it satisfies the definition of \emph{suddenly appear} at that position. However, when it leaves that position, it satisfies the definition of \emph{suddenly disappear}.

According to the definition, if the time of the first observation of an object is later than the time of the first observation of the ground below it, then the object is considered as a \emph{suddenly appear} dynamic object. On the contrary, if the time of the last observation of an object is earlier than the time of the last observation of the ground, then the object is considered as a \emph{suddenly disappear} dynamic object. We refer to this method of determining dynamic objects as observation time difference.

\section{Methodology}
\label{sec::methodology}
The overall framework of our proposed dynamic objects removal method is illustrated in Fig. \ref{fig::pipeline}, which includes three modules: ground segmentation, map management and dynamic removal.

\subsection{Ground Segmentation}
Since our proposed dynamic removal approach needs to constantly compare the observation time difference between ground and non-ground objects, the accuracy of ground segmentation will affect the accuracy of dynamic removal. Therefore, we have adopted a two-step ground segmentation process, whereby candidate ground points are first extracted through preprocessing, and then they are refined to obtain the final ground point set.

In the preprocessing, we refer to the method proposed in\cite{bogoslavskyi2016fast} to remove most of the non-ground points. Then, we introduce PCA proposed in\cite{zermas2017fast} to further refine it and obtain the final ground point set $\mathbf{G}$, while the remaining points are classified as non-ground point set $\mathbf{U}$. Since we removed most of the non-ground points in the preprocessing, the ground fitting process will be relatively more accurate. This two-step ground segmentation method aims to minimize the number of dynamic points in the segmented ground as much as possible, which is crucial for improving the effectiveness of dynamic removal.

\subsection{Map Management}
It is difficult to accurately identify dynamic \emph{object} in point cloud maps, so we use voxels\cite{koide2021voxelized}, \cite{bai2022faster} to represent \emph{object}. If a voxel is detected to be dynamic, all points in the voxel are classified as dynamic points.

Therefore, following \cite{bai2022faster}, we build a voxel map to manage the point cloud. In order to facilitate the management of ground points, non-ground points and dynamic points, we divide the global map into three parts: ground submap $^{g}\mathcal{V}$, non-ground submap $^{n}\mathcal{V}$ and dynamic submap $^{d}\mathcal{V}$. These three submaps share the same coordinate system, but are stored separately. 

In addition, in order to obtain the observation time of each voxel, we store the index of LiDAR frame for each point besides storing the points in the voxel, and record all the frame indexes in a Set. Then, the smallest frame index in the set can represent the time when the voxel was first observed, while the largest frame index can represent the time when the voxel was last observed. For convenience, we will refer to the voxels located in the ground submap, non-ground submap, and dynamic submap as ground voxels, non-ground voxels, and dynamic voxels respectively in the following paper.

\subsection{Dynamic Removal}

Based on the observation time difference we introduced in Section \ref{sec::classification}, we can compare the observation time difference between each non-ground voxel and ground voxel below it to determine whether the voxel is dynamic or not. Indeed, comparing all voxels with the ground for each iteration is impractical for online processing. Therefore, we propose downward retrieval and upward retrieval methods to remove the \emph{suddenly appear} and \emph{suddenly disappear} dynamic voxels respectively. In addition, there will always be some misidentified static voxels anyway, so we added static restoration process.

Using the input pose $\mathbf{T}^{W}$ (usually comes from the SLAM front-end), we can transform $\mathbf{G}$ and $\mathbf{U}$ into the world frame and designate them as $\mathbf{G}^{W}$ and $\mathbf{U}^{W}$, respectively.

\subsubsection{Downward Retrieval}
\label{subsubsec::down_ret}

\begin{algorithm}[t]  
    \SetAlgoLined
    \For{ $\mathbf{p}_{i}^{W} \in \mathbf{U}^{W}$}{
         Let $^{n}\mathbf{V}_{i}$ be the non-ground voxel of $\mathbf{p}_{i}^{W}$; \\
         Let $^{n}\gamma_{min}$ be the minimum frame index of $^{n}\mathbf{V}_{i}$; \\
         \For{k = 1, 2, ...}{
             $^{g}\mathbf{V}_{i}$ = $^{n}\mathbf{V}_{i}$ - $k \cdot (0, 0, 1)^T$; \\
             \If{$^{g}\mathbf{V}_{i} \neq \emptyset$ }{
                Let$^{g}\gamma_{min}$ be the minimum frame index of $^{g}\mathbf{V}_{i}$; \\
                \textbf{break};
            }
        }
        \If{$^{n}\gamma_{min} - ^{g}\gamma_{min} > \tau_{ret}$}{
             Find dynamic voxel $^{d}\mathbf{V}_{i}$ of $\mathbf{p}_{i}^{W}$; \\
             $^{d}\mathbf{V}_{i}$ = $^{d}\mathbf{V}_{i} \cup {^{n}\mathbf{V}_{i}}$; \\
             $^{n}\mathbf{V}_{i} = \emptyset$; \\
        }
    }
    \caption{Downward Retrieval}
    \label{alg::downward}
\end{algorithm}

The downward retrieval process we proposed is shown in Algorithm \ref{alg::downward}. Since the \emph{suddenly appear} dynamic objects are the objects that can be observed at current time, it must exist in $\mathbf{U}^{W}$. Therefore, to achieve the removal of \emph{suddenly appear} dynamic objects, we only need to examine the voxel containing $\mathbf{U}^{W}$.

 For each point $\mathbf{p}_{i}^{W}$ in $\mathbf{U}^{W}$, we calculate its non-ground voxel $^{n}\mathbf{V}_{i}$ in non-ground submap. Then, we count all frame indexes in $^{n}\mathbf{V}_{i}$, and designate the smallest frame index as $^{n}\gamma_{min}$. $^{n}\gamma_{min}$ can represent the time when $^{n}\mathbf{V}_{i}$ is first observed.

Next, in the ground submap, we retrieve the ground voxel $^{g}\mathbf{V}_{i}$ within 3 meters by descending along $z$-axis from $^{n}\mathbf{V}_{i}$. Finally, we count all the frame indexes in $^{g}\mathbf{V}_{i}$, and designate the smallest frame index as $^{g}\gamma_{min}$. $^{g}\gamma_{min}$ can represent the time when the ground voxel is first observed.

It can be demonstrated that $^{n}\mathbf{V}_{i}$ is a \emph{suddenly appear} dynamic voxel if the following criteria are satisfied: 
\begin{equation}
\label{eq::down_ret}
    ^{n}\gamma_{min} - {^{g}\gamma_{min}} > \tau_{ret}
\end{equation}
where $\tau_{ret}$ is a pre-defined threshold. Then, we move all points as well as all frame indexes in $^{n}\mathbf{V}_{i}$ into dynamic submap. 


\begin{figure}[h]
\centering
\includegraphics[width=0.95\linewidth]{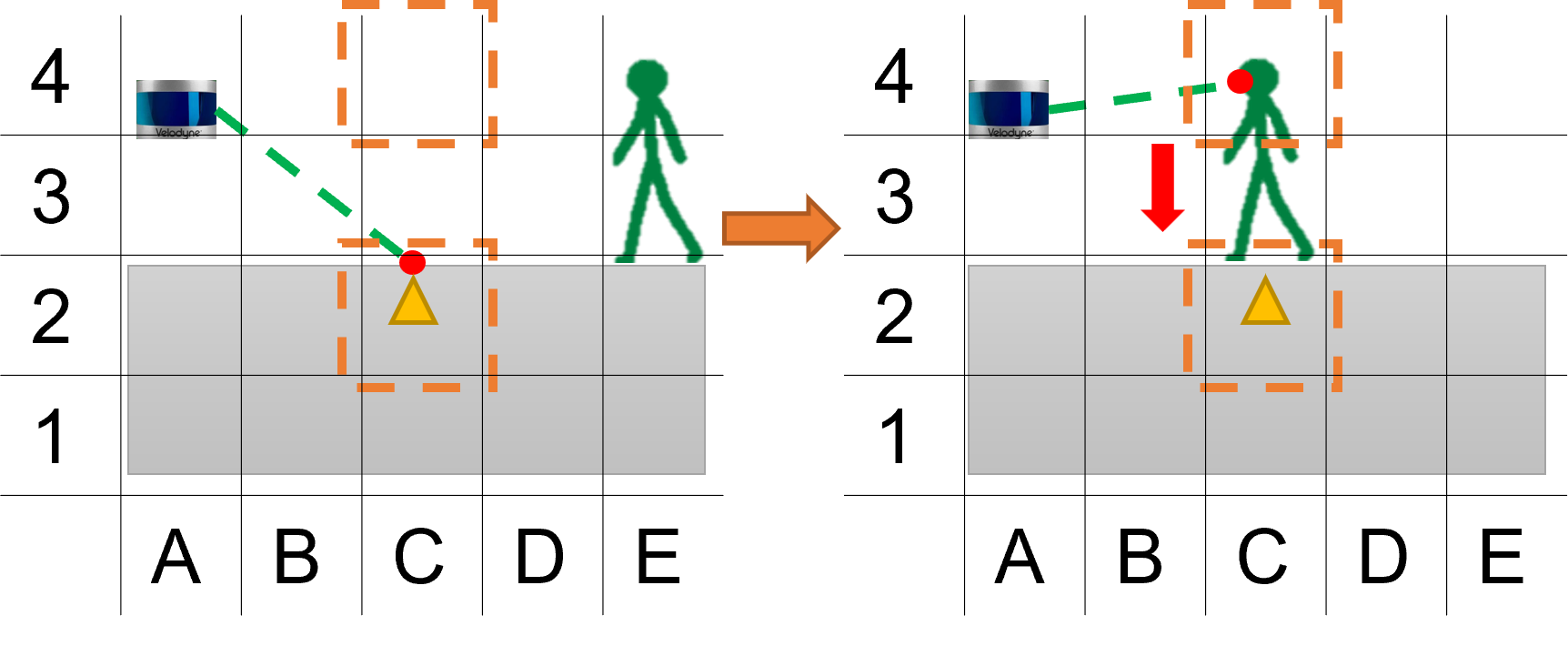}
\caption{Downward retrieval. The image on the left represents the $i$-th LiDAR frame, while the image on the right represents the $j$-th LiDAR frame ($j > i$). The points observed by LiDAR are represented by red dots. The object was at position E at the $i$-th LiDAR frame, and it moved to position C at the $j$-th LiDAR frame. We first observed the non-ground voxel (C, 4) in the $j$-th LiDAR frame, but we had already observed the ground voxel (C, 2) below it in the $i$-th frame. Therefore, we compare the minimum LiDAR frame index between these two voxels to determine whether (C, 4) is dynamic or not.}
\label{fig::retrieval_down}
\end{figure}

As shown in Fig. \ref{fig::retrieval_down}, the non-ground voxel (C, 4) was first observed in the $j$-th frame. Therefore, the minimum LiDAR frame index for (C, 4) is recorded as $j$. Then, we find the ground voxel (C, 2) by descending along $z$-axis from (C, 4). Since (C, 2) was observed in the $i$-th frame, the minimum LiDAR frame index for (C, 2) is $i$. According to Eq. \ref{eq::down_ret}, if $j-i > \tau_{ret}$, then the non-ground voxel (C, 4) is considered as a dynamic voxel.

\subsubsection{Upward Retrieval}
Since the \emph{suddenly disappear} dynamic objects can no longer be observed at current time, that is, they do not exist in $\mathbf{U}^{W}$. Indeed, the ground below them can still be observed. Therefore, we use points in $\mathbf{G}^{W}$ to retrieve upwards to find  \emph{suddenly disappear} dynamic objects.

For each point $\mathbf{p}_{i}^{W}$ in $\mathbf{G}^{W}$, we calculate its corresponding voxel $^{g}\mathbf{V}_{i}$ in the ground submap. Then, we define the largest frame index in $^{g}\mathbf{V}_{i}$ as $^{g}\gamma_{max}$. Next, retrieve all the non-ground voxels within 3 meters by ascending along the $z$-axis in the non-ground submap. We check each of these non-ground voxels $^{n}\mathbf{V}_{i}$ in turn, and define the largest frame index in it as $^{n}\gamma_{max}$.

Finally, if the following criteria is satisfied: 
\begin{equation}
    ^{g}\gamma_{max} - {^{n}\gamma_{max}} > \tau_{ret}
\end{equation}
It can be demonstrated that $^{n}\mathbf{V}_{i}$ is a \emph{suddenly disappear} dynamic voxel. 

\subsubsection{Static restoration}
In Section \ref{sec::classification}, we proposed the assumption that static objects should appear and disappear simultaneously with the ground. However, in real-world scenarios, this assumption does not always hold true due to the influence of ground slope and LiDAR measurement noise. This inevitably leads to some static voxels being mistakenly identified as dynamic voxels. Therefore, we introduce a static restoration module to place these misclassified static voxels back into the non-ground submap. We assume that if the total times of observation of a non-ground voxel and the ground voxel is similar, then the voxel should be static. Based on this, we designed the process of static restoration.

We find the dynamic voxel $^{d}\mathbf{V}_{i}$ in the dynamic submap during the downward retrieval and define the total number of frame indexes in $^{d}\mathbf{V}_{i}$ as $^{d}\gamma_{sum}$. $^{d}\gamma_{sum}$ can represent the total times the voxel has been observed. Correspondingly, we define the total number of frame indexes in the ground voxel $^{g}\mathbf{V}_{i}$ as $^{g}\gamma_{sum}$.

If the difference between $^{g}\gamma_{sum}$ and $^{d}\gamma_{sum}$ is less than a pre-defined threshold $\tau_{res}$, $^{d}\mathbf{V}_{i}$ is proved to be a static voxel that has been misidentified. Thus, we move all points and all frame indexes in $^{d}\mathbf{V}_{i}$ back into the non-ground submap.

\begin{table*}[h]
\caption{Comparison results of dynamic removal using SemanticKITTI dataset.  Metrics are expressed as PR[\%]/RR[\%]/$\mathbf{F}_{1}$ score.}
\label{tab::KITTIdr}
\resizebox{\textwidth}{!}{%
\begin{tabular}{lccccc}
    \toprule
    Method & Seq.00 & Seq.01 & Seq.02 & Seq.05 & Seq.07 \\ 
    \midrule
    Removert-RM3\cite{kim2020remove} & 85.502/\textbf{99.354}/0.919 & 94.221/93.608/0.939 & 76.319/96.799/0.853 & 86.900/87.880/0.874 & 80.689/98.822/0.888 \\
    Removert-RM3+RV1\cite{kim2020remove} & 86.829/90.617/0.887 & 95.815/57.077/0.715 & 83.293/88.371/0.858 & 88.170/79.981/0.839 & 82.038/95.504/0.883 \\
    ERASOR-0.2\cite{lim2021erasor} & 93.980/97.081/0.955 & 91.487/95.383/0.934 & 87.731/97.008/0.921 & 88.730/98.262/0.933 & 90.624/\textbf{99.271}/0.948 \\
    ERASOR-0.4\cite{lim2021erasor} & 80.769/98.353/0.887 & 79.807/95.080/0.868 & 78.319/98.464/0.872 & 72.440/97.831/0.832 & 82.875/98.221/0.899 \\
    ERASOR2\cite{lim2023erasor2} & 98.788/98.582/0.987 & 96.879/94.629/0.957 & 98.523/\textbf{99.709}/\textbf{0.991} & \textbf{97.582}/98.992/0.983 & \textbf{98.977}/98.459/\textbf{0.987} \\
    \midrule
    DynamicFilter\cite{fan2022dynamicfilter} & 90.070/91.090/0.906 & 87.950/87.690/0.878 & 88.020/86.100/0.871 & 90.170/84.650/0.873 & 87.940/86.800/0.874 \\
    Nonparametric\cite{park2022nonparametric} & 94.050/97.190/0.956 & 91.815/94.096/0.929 & 91.208/95.510/0.933 & 93.820/95.740/0.947 & 91.146/97.284/0.941 \\
    Ours & \textbf{98.819}/98.686/\textbf{0.988} & \textbf{99.013}/\textbf{95.494}/\textbf{0.972} & \textbf{98.682}/98.954/0.988 & 97.316/\textbf{99.520}/\textbf{0.984} & 96.490/98.492/0.975 \\
    \bottomrule
\end{tabular}%
}
\end{table*}

\section{Experiments} \label{sec::experiments}
In this section, in order to prove the effectiveness of our proposed method, we conducted online dynamic object removal experiments on the public SemanticKITTI dataset and challenging dataset. Qualitative and quantitative comparisons with state-of-the-art algorithms are conducted to evaluate our proposed method. In order to fairly test the performance of our algorithm, all experiments were performed on the same laptop with an AMD R7-5800H CPU and 16GB RAM.

\subsection{Evaluation Metrics}
We use $preservation$ $rate$ (PR) and $rejection$ $rate$ (RR) proposed in \cite{lim2021erasor} and their harmonic mean $\mathbf{F}_{1}$ score to evaluate performance of dynamic removal. PR and RR represent the preservation rate of static objects and the rejection rate of dynamic objects respectively, and their definitions are as follows:
\begin{itemize}
	\item PR: $\frac{\mbox{\# of preserved static points}}{\mbox{\# of total static points on the raw map}}$
	\item RR: $1 - \frac{\mbox{\# of preserved dynamic points}}{\mbox{\# of total dynamic points on the raw map}}$
\end{itemize}
Since PR and RR are calculated voxel-wise, we set the voxel size to 0.2 for all methods for a fair comparis.

\subsection{Comparison in SemanticKITTI Datasets}
We compare the proposed method with other state-of-the-art dynamic removal methods, including Removert\cite{kim2020remove}, ERASOR\cite{lim2021erasor} and ERASOR2\cite{lim2023erasor2} for offline removal, and DynamicFilter\cite{fan2022dynamicfilter} and Nonparametric\cite{park2022nonparametric} for online removal. The poses required by our method are provided by SuMa\cite{behley2018efficient} which contains inherent uncertainty. Additionally, since dynamic objects are not always present in SemanticKITTI, referring to \cite{lim2021erasor}, we select frames 00 (4,390 - 4,530), 01 (150 - 250), 02 (860 - 950), 05 (2,350 - 2,670), and 07 (630 - 820) as dynamic removal benchmark where the numbers in parenthesis indicate the start and end frames.  

In all segments, we set identical dynamic removal parameters, where $\tau_{ret} = 7$ and $\tau_{res} = 15$. Table \ref{tab::KITTIdr} shows the experimental results of ours and other methods on five segments of SemanticKITTI. In Table \ref{tab::KITTIdr}, RM3 denotes the results after three remove stages with per-pixel resolutions of $0.5^\circ$, $0.45^\circ$ and $0.4^\circ$ in Removert\cite{kim2020remove}. RM3+RV1 means the result of RM3 followed by a revert stage with the resolution per pixel of $0.55^\circ$. ERASOR-0.2 and ERASOR-0.4 denote the different sizes of scan ratio threshold in ERASOR\cite{lim2021erasor}. 

As shown in Table \ref{tab::KITTIdr}, our method achieves comparable performance to ERASOR2 and outperformed other methods in semanticKITTI. By introducing instance segmentation, ERASOR2 rejects dynamic points at an instance-level. Therefore, ERASOR2 outperforms our method in some segments. Particularly, in the Seq.02 segment, it achieves an impressive $\mathbf{F}_{1}$ score of 0.99. However, ERASOR2 is an offline algorithm that requires a prior map and has a higher time consumption, making it unsuitable for online processing. As for other methods, Removert-RM3 is too confident in removing dynamic objects, resulting in a low PR. In the revert process, some dynamic points will be put back into the map, so the RR of Removert-RM3+RV1 is the lowest. The score of ERASOR in PR is also unsatisfactory, especially after the scanning ratio threshold is increased, the PR of ERASOR-0.4 has dropped sharply. Compared with DynamicFilter and Nonparametric, which are online dynamic removal methods, our method achieves better results because it is not limited by the information of map. It should be noted that our method is stable across all 5 segments while other on-line methods vary greatly in performance across different segments, which means that our method is more robust than other methods.

\begin{figure*}[h]
    \centering
    \subfigure[Original Map]{
        \includegraphics[width=0.23\linewidth]{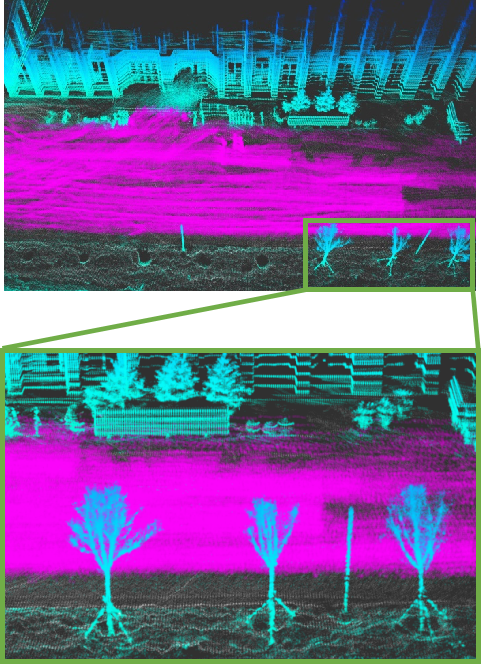}
        }
    \subfigure[Removert]{
        \includegraphics[width=0.23\linewidth]{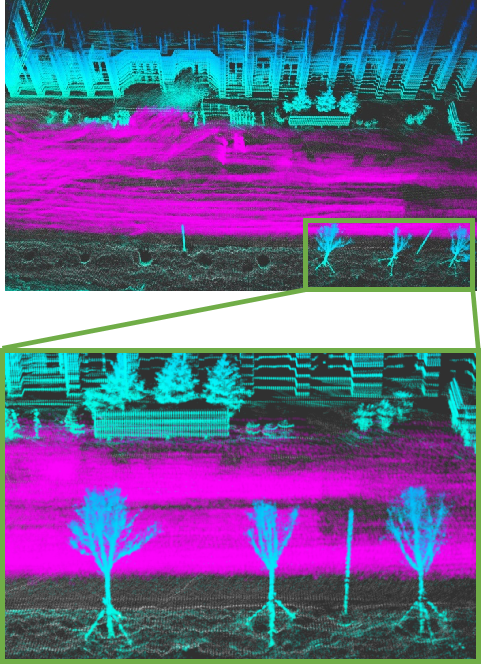}
        }
    \subfigure[ERASOR]{
        \includegraphics[width=0.23\linewidth]{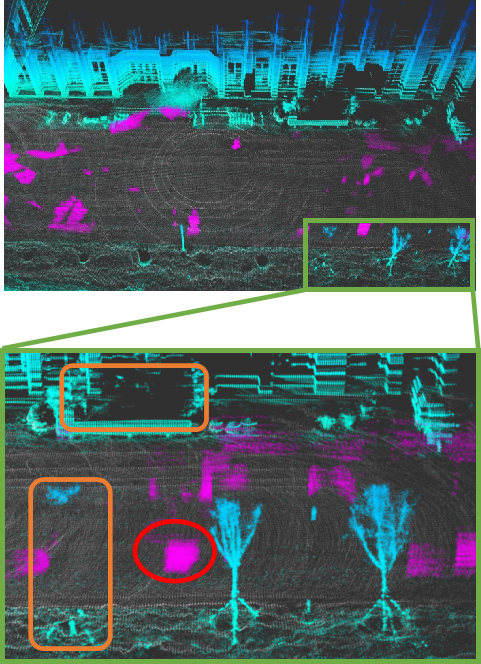}
        }
    \subfigure[Ours]{
        \includegraphics[width=0.23\linewidth]{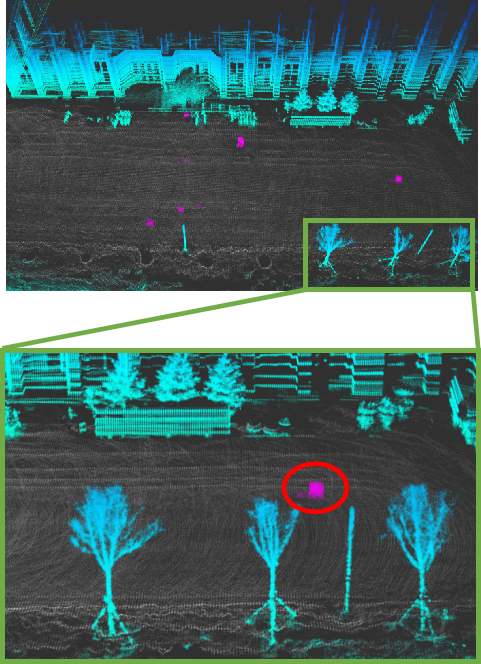}
        }
\caption{The dynamic removal results in challenging scenarios, the partial enlarged image is shown below. The purple points in the picture is traces left by dynamic objects, and the area with serious error removal is in orange boxes.}
\label{fig::dingxing}
\end{figure*}

\subsection{Comparison in challenging scenarios}
To further validate the algorithm's robustness, we conducted experiments on dynamic object removal in challenging scenarios with high pedestrian density. As of now, ERASOR2, DynamicFilter, and Nonparametric algorithms have not been open-source, so we only compared our method with ERASOR and Removert, and the comparative results are shown in Fig. \ref{fig::dingxing}. The selected scene is the road in front of the school cafeteria, and the data recording time coincides with the students' dismissal time. At this time, almost all students who finish classes head to the cafeteria for meals, leaving behind a significant amount of traces in the image. As shown in Fig. \ref{fig::dingxing}(a), the middle area of the image is almost filled with dynamic points.

The dynamic object removal effectiveness of Removert in such a crowded scene is limited. While Removert managed to remove some dynamic objects, there are still numerous traces left in the image attributable to the presence of dynamic objects. The results of ERASOR are shown in Fig. \ref{fig::dingxing}(c), where most of the dynamic traces have been removed, but there are still some traces remaining. Additionally, ERASOR mistakenly removed a significant number of static objects, such as the wall and tree trunk highlighted in the orange boxes in the image.

By comparing the observation time difference between the objects and the ground, our method achieved a more significant dynamic object removal effect. As shown in Fig. \ref{fig::dingxing}(d), our algorithm successfully removed almost all dynamic objects and preserved important details such as tree trunks and buildings. The remaining purple points in the image are caused by objects that remained stationary throughout, so our algorithm did not consider it as a dynamic object.

\begin{table}[t]
    \centering
    \caption{Runtime Per Iteration of SemanticKITTI Dataset.}
    \label{tab::KITTIrt}
    \resizebox{0.7\linewidth}{!}{%
        \begin{tabular}{lc}
            \toprule
                Method & Runtime/iteration [ms] \\
            \midrule
                Removert\cite{kim2020remove} & 73.4 \\
                ERASOR\cite{lim2021erasor} & 72.6 \\
                Nonparametric\cite{park2022nonparametric} & 59.9 \\
                Ours & \textbf{23.8} \\
            \bottomrule
        \end{tabular}%
    }
\end{table}

\begin{table}[t]
    \centering
    \caption{ Each Module's Runtime of Our Method.}
    \label{tab::KITTIet}
    \resizebox{0.7\linewidth}{!}{%
        \begin{tabular}{lc}
            \toprule
                Module & Runtime/iteration [ms] \\
            \midrule
                Ground Segmentation & 13.0 \\
                Map Management & 5.3 \\
                Dynamic Removal & 5.5 \\
                Total & {23.8} \\
            \bottomrule
        \end{tabular}%
    }
\end{table}

\subsection{Algorithm Speed}
In our algorithm, no matter downward retrieval or upward retrieval, it only needs to compare the observation time of voxels without additional calculation. Moreover, the number of voxels to be retrieved in each frame is much smaller than the number of points. Therefore, compared to other state-of-the-art algorithms, the computing time of our method is reduced by at least 60$\%$, as shown in the Table \ref{tab::KITTIrt}. 

In addition, in our method, the dynamic removal modules can complete the task within only several milliseconds, and more computing resources are actually used in other modules, as shown in the Table \ref{tab::KITTIet}. The ground segmentation module takes the longest time, up to 13 milliseconds, because we have adopted a two-step ground segmentation algorithm. However, this two-step ground segmentation algorithm minimizes the number of dynamic points in the ground as much as possible, which greatly improves the performance of the dynamic removal.

\section{Conclusion} 
\label{sec::conclusion}
In this paper, we propose a new method for online dynamic objects removal for ground vehicles. Our method identifies dynamic objects by comparing the observation time difference between objects and the ground. Our method achieves excellent dynamic removal performance by only using the map at the current moment, and is more robust than other methods. But since our method relies heavily on the comparison with the ground, it will perform poorly in uneven terrain scenarios and unable to detect aerial objects. 

\addtolength{\textheight}{-9cm}   

\bibliographystyle{IEEEtran}
\bibliography{reference}

\end{document}